\documentclass[preprint,12pt]{elsarticle}
\usepackage{lmodern}

\usepackage{amssymb}
\usepackage{tabularx}
\usepackage{amsmath}
\usepackage{float} 
\usepackage{array}
\usepackage{calc}

\journal{Engineering Reports}

\begin{document}

\begin{frontmatter}


\title{Multi-Objective Neural Network-Assisted Design Optimization of Soft Fin-Ray Fingers for Enhanced Grasping Performance}

\author[label1]{Ali Ghanizadeh}

\affiliation[label1]{organization={School of Mechanical Engineering, College of Engineering, University of Tehran, Tehran, Iran}}

\title{}
\author[label1]{Ali Ahmadi}
\author{Arash Bahrami\corref{cor1}\fnref{label1}}
\ead{arash.bahrami@ut.ac.ir}
\cortext[cor1]{Corresponding author.}

\begin{abstract}
The internal structure of the Fin-Ray fingers plays a significant role in their adaptability and grasping performance. However, modeling the grasp force and deformation behavior for design purposes is challenging. When the Fin-Ray finger becomes more rigid and capable of exerting higher forces, it becomes less delicate in handling objects. The contrast between these two gives rise to a multi-objective optimization problem. We employ the finite element method to estimate the deflections and contact forces of the Fin-Ray fingers grasping cylindrical objects, generating a dataset of 120 simulations. This dataset includes three input variables: the thickness of the front and support beams, the thickness of the crossbeams, and the equal spacing between the crossbeams, which are the design variables in the optimization. This dataset is then used to construct a multilayer perceptron (MLP) with four output neurons predicting the contact force and tip displacement in two directions. The magnitudes of maximum contact force and maximum tip displacement are two optimization objectives, showing the trade-off between force and delicate manipulation. The set of solutions is found using the non-dominated sorting genetic algorithm (NSGA-II). The results of the simulations demonstrate that the proposed methodology can be used to improve the design and grasping performance of soft grippers, aiding to choose a design not only for delicate grasping but also for high-force applications.
\end{abstract}


\end{frontmatter}



\section{Introduction}
\label{sec1}

Robotic grippers used to be made of rigid parts and links in most cases, but recent progress in soft robotics and related fields has attracted much attention to soft grippers and boosted their development. Soft grippers have the ability to handle a wider range of objects compared to their rigid counterparts, while also allowing for the use of simpler control frameworks \cite{article}.

Fin-Ray grippers are a type of soft gripper, and their basic concept was biologically inspired by Leif Kniese's observation of a phenomenon in fish. By applying force to the structure of the fish's fins, it bends in the opposite direction of the force. By following up on this observation and studying the structure in collaboration with Rudolf Bannasch, the Fin-Ray effect was introduced \cite{pfaff2011application}. Fin-Ray grippers can be used to grasp different delicate and sensitive objects such as eggs \cite{app11093858} and fruits \cite{AN2025110118}.

A key factor influencing the performance of Fin-Ray grippers is the finger design, and several studies focused on introducing new designs or improving existing ones based on the desired task. In a study, an optimized gripper based on the Fin-Ray effect with an integrated linkage mechanism was designed to grasp and harvest tomatoes. Their design provided a balanced force distribution on tomatoes and also included a blade at the tip of the Fin-Ray finger that automatically cuts the stems while harvesting \cite{AN2025110118}. Crooks et al. designed a novel gripper inspired by the traditional Fin-Ray grippers. However, their gripper was activated via a motor-tendon mechanism and made of a combination of soft and hard materials, all printable as a complete gripper. Hard parts included crossbeams, supports, and fingernails. They compared their optimized design with the common Fin-Ray grippers both experimentally and by simulation. As a result of their design, the gripper's tip and structure could move more and perform better gripping, showing 15\% greater deformation compared to a regular Fin-Ray finger. In addition, the contact area was larger, resulting in a more stable grip, and the gripper was able to handle about 40\% heavier loads \cite{10.3389/frobt.2016.00070}. 

Besides the Fin-Ray finger design, the material from which they are made plays an important role in their performance. Hence, a study was conducted on material selection for Fin-Ray grippers. They simulated the response of a single gripper finger to a static force in ANSYS Workbench. Their objective was to choose the best material according to their design goals \cite{8813388}.

To optimize the design of Fin-Ray fingers, several geometric factors can be considered, including rib (crossbeam) thickness, the crossbeam angles, and the angle increment or spacing distance between crossbeams. The optimization problem is carried out to achieve enhanced shape adaptation and force generation, reduced initial contact forces, or improved performance during the layer jamming phase \cite{9115969, 10.3389/frobt.2020.590076}. However, the optimization variables are not always limited to these, since the finger's internal structure can also be directly studied. Topology optimization is considered a way to improve the grasping ability and safety factor of Fin-Ray fingers. Lakshmi Srinivas et al. compared fingers with different internal layouts both with and without topology optimization. Comparing the best layouts of two groups, while the non-optimized finger offered the best wrapping ability, the optimal one provided a superior balance of wrapping ability, structural strength, and lightweight design, making it an efficient choice for high-load and versatile grasping tasks \cite{article12}.  Yao et al. investigated the internal crossbeam structures of Fin-Ray grippers. They considered four different layouts, including one without crossbeams and used FEM to enhance the adaptability of the Fin-Ray fingers. They found that removing the internal structure enhanced the finger's ability to conform to delicate items while minimizing applied force \cite{Yao_Fang_Li_2023}.

FEM is widely considered an accurate method, capable of simulating and modeling linear and multiphysics problems. In the context of soft robotics, it provides the ability to simulate complex structures and nonlinear elastic materials, two essential aspects of understanding soft robotics. However, when dealing with these problems, FEM is computationally expensive and may face convergence difficulties. Combining data-driven methods such as Machine Learning (ML) techniques with FEM offers new opportunities in the modeling and development of soft robotic systems. This combination creates a powerful tool for optimizing soft robotics structures, material selection, and control strategies \cite{Jin_2025}. 

De Barrie et al. developed a neural network for real-time prediction of contact forces and stress maps in soft Fin-Ray grippers. They prepared a dataset using FEM simulations of the gripper that interacts with different grasped objects, varying in size, shape, height, and angle of approach. Their network demonstrated promising results and was able to predict cases that were not in the training data. However, there were limitations in the real-world applications, including that the network's performance dropped as the camera angle increased \cite{10.3389/frobt.2021.631371}. Yao et al. proposed a two dimensional kinetostatic model of a soft Fin-Ray finger. The model was able to calculate the total contact force and deformations, and it was found out that the results obtained within their model exhibited similar outputs and accuracy compared to those of the FEM simulation. However, friction was not considered in the model. Next, they used this model to optimize the finger structure \cite{YAO2024105472}. Also, in another study, a kinetostatic model was proposed for multi-crossbeam soft fingers to accurately estimate contact forces during object grasping. Their approach not only enabled efficient force prediction but also incorporated the influence of varying stiffness across different finger segments, providing valuable insights for optimizing the design and performance \cite{doi:10.1177/0278364920913926}. Ghanizadeh et al. estimated the contact forces in soft Fin-Ray grippers for grasping cylindrical objects using FEM and experimental validation. However, this study considered only one internal structure for the Fin-Ray finger which limits the generalization of the findings \cite{10903660}. Xu et al. introduced a compliant adaptive Fin-Ray gripper. They changed the traditional Fin-Ray gripper by putting rigid parts into it, making its force-deformation behavior linear. Then, they trained a neural network using data from the finite element method (FEM). This made it possible to calculate external force from the finger's deformation. However, the accuracy of the estimated force dropped as the applied force point moved from the middle to the finger's ends \cite{9380238}.  Wang et al. proposed a physics-informed neural network (PINN) to model a soft Fin-Ray structure, where the minimum potential energy was integrated into the loss function based on elasticity theory. They trained two models, with and without the data from a real Fin-Ray finger. Their experiments showed that the PINN without the real data and the FEM had nearly the same accuracy, while the PINN with the real data stood out and improved the accuracy \cite{10802217}.

The importance of optimizing the Fin-Ray finger structure has been widely recognized. Since FEM simulations are computationally expensive, this study aims to employ a neural network that mimics the behavior of a Fin-Ray finger and is trained on a small dataset obtained from FEM simulations. In this work, we use a population size of 500 and 100 generations when applying NSGA-II in the optimization process. This would require about 50,000 FEM runs, which is prohibitively expensive. Instead, we train the MLP by running FEM simulations for only 120 Fin-Ray fingers, enabling fast and accurate evaluation. Next, the MLP is used in a multi-objective optimization problem to identify the Fin-Ray finger designs. While surrogate-based approaches have been explored in some studies on Fin-Ray grippers, applying a neural network as the main evaluation model for multi-objective optimization of internal structural parameters has not been studied, and this work contributes to filling that gap. This approach demonstrates the feasibility of achieving good performance with a limited dataset while significantly reducing computational cost and time. In the final section, we provide the conclusion, summarizing the key results of the article.


\renewcommand{\arraystretch}{1.3}
\setlength{\arrayrulewidth}{0.2mm}
\setlength{\tabcolsep}{18pt}
\section{Finite Element Method}
In this section, we examine the Fin-Ray finger contact force and tip displacement during interaction with a 20 mm diameter cylinder. We use the data from this part to train the multilayer perceptron model in the following phases of our study.

\begin{figure} [H]
    \centering
    \includegraphics[scale=1]{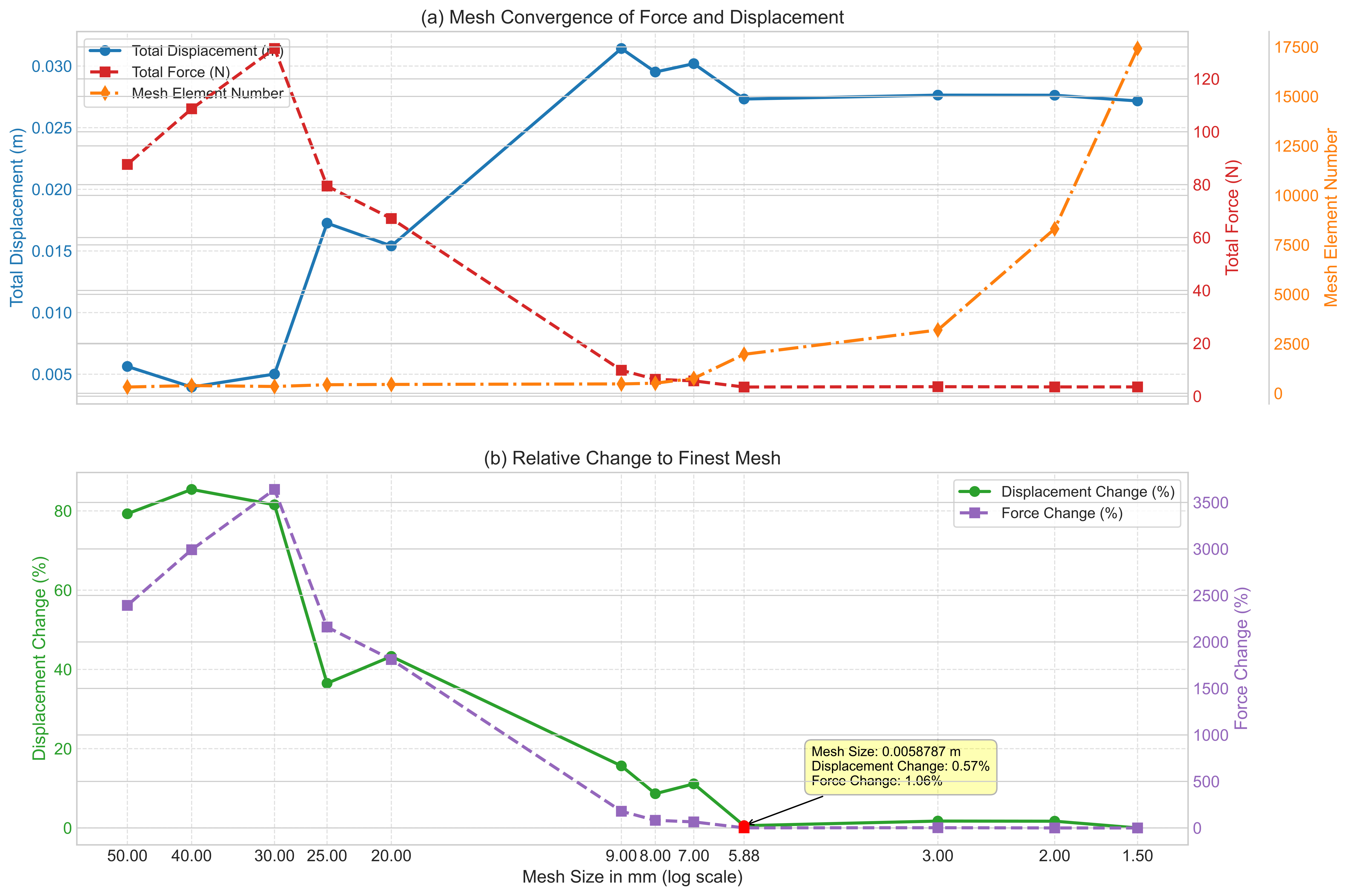}
    \caption{Mesh convergence study: (a) Total displacement, force and number of elements and (b) Relative changes with respect to the finest mesh values.}
    \label{meshDependency}
\end{figure}

We use FEM to estimate the Fin-Ray finger contact force and tip displacement for different internal structures. Static structural analysis is employed to analyze contact forces with the ANSYS software. The cylindrical object with a diameter of 20 mm contacts the Fin-Ray finger at the midpoint of the front beam. This cylinder is considered to be linearly elastic with the properties of the acrylonitrile butadiene styrene (ABS) material found in manufacturing technical specifications. Thermoplastic polyurethane (TPU) is widely employed in soft grippers due to its flexibility and resilience \cite{Xin2023TheRO}. In this study, all simulated Fin-Ray fingers have the same material of TPU 95A, which is not included in the standard ANSYS material library \cite{10.3389/frobt.2021.631371}. Hence, we consider Young's modulus of 26 MPa and Poisson's ratio of 0.48 as the mechanical properties of this material \cite{Yao_Fang_Li_2023}. It is assumed that the Fin-Ray finger experiences a 2.5 cm base displacement in the direction of the cylindrical object. Afterwards, we define the grasp frictional contact region and additional essential constraints for the simulations, such as the fixed displacement constraints. The friction coefficient is assumed to be 0.25 and the contact is defined as "frictional" in ANSYS simulations \cite{Yao_Fang_Li_2023}. We employ the default ANSYS mechanically controlled meshing algorithm, resulting in an average element size of 5.88 mm. The default ANSYS convergence criteria for force and displacement are used. SOLID186 elements are applied to 3D solid regions, and CONTA174 elements are used to model contact and sliding interactions between soft deformable surfaces and 3D solids. The total element count varies among different Fin-Ray configurations. In addition, we conduct a convergence study to make sure that our numerical results are independent of the mesh. A mesh dependency study for one of the Fin-Ray structures with 1.5 mm front and support beam thicknesses, 0.6 mm crossbeam thickness, and 10 mm spacing between crossbeams is presented in Fig.~\ref{meshDependency}. It can be seen that, with errors of approximately 0.57\% and 1.06\% in total displacement and total force relative to the finest mesh, the selected mesh size of 5.88 mm demonstrates mesh independence.

We perform FEM simulations for Fin-ray fingers with different internal structures to determine the tip displacement and estimate the contact force for each case. To this end, we change three design parameters of the Fin-Ray finger internal structure: thickness of the front and support beams, the thickness of the crossbeams, and the equal spacing between each crossbeam. The thicknesses of the support and front beams are considered equal in this study. These parameters are vital because of their importance in influencing our target features. The range in which these parameters are varied is summarized in Table 1. This range can be selected differently depending on the application for which the Fin-Ray finger is used. We simulate all possible structures within this feasible range of thicknesses. As can be seen there are \(6 \times 5 \times 4 = 120\) possible combinations and a total of 120 distinct structures is successfully simulated and included in the dataset. Other internal structural parameters, such as the angle of inclination of the crossbeams and the length of the finger base, are remained unchanged.

\setlength{\tabcolsep}{3pt}
\renewcommand{\arraystretch}{1.1}

\begin{table}[H]
\begin{center}
\caption{Values for each design parameter of Fin-Ray finger}
\label{tab1}
    {\small
    \resizebox{0.85\columnwidth}{!}{%
    \begin{tabular}{|c | c | c |}
    \hline
    \textbf{\textit{Parameter Name}}& \textbf{\textit{Range}}& \textbf{\textit{Increment Size}} \\
    \hline
    Thickness of front and support beams& 1.5-4 mm& 0.5 mm \\
    \hline
    Thickness of crossbeams& 0.8-1.6 mm& 0.2 mm \\
    \hline
    Spacing between each crossbeam& 10-16 mm& 2 mm \\
    \hline
    \end{tabular}%
    }
}
\end{center}
\end{table}

\begin{figure}[H]
    \centering
    \includegraphics[scale=3]{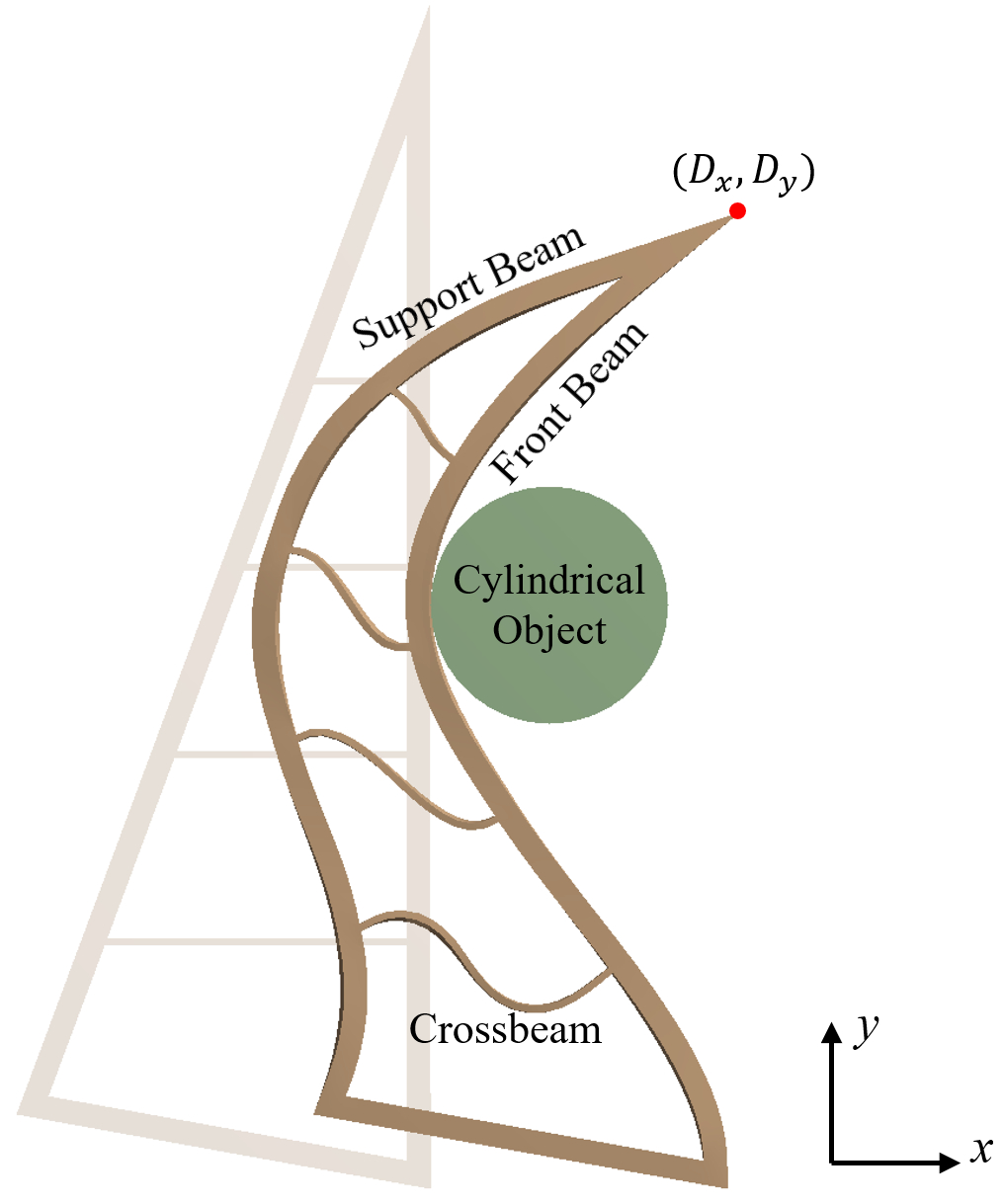}
    \caption{A schematic of the Fin-Ray finger in simulation}
    \label{fig}
\end{figure}

The front beam and crossbeams are parallel to the \textit{y} and \textit{x} axes just before contact, respectively. An illustration of the Fin-Ray finger before and after contact with the cylindrical object is shown in Fig.~\ref{fig}, where $D_x$ and $D_y$ are the tip displacements of the Fin-Ray finger in the \textit{x}- and \textit{y}-directions. The crossbeams, front beam and support beams are also displayed in Fig.~\ref{fig}. Furthermore, the base length, the front and support beam lengths, and the width of the Fin-Ray finger are assumed to be 35 mm, 100 mm, and 30 mm, respectively.

\begin{figure}[H]
 \centering
 \includegraphics[scale=0.17]{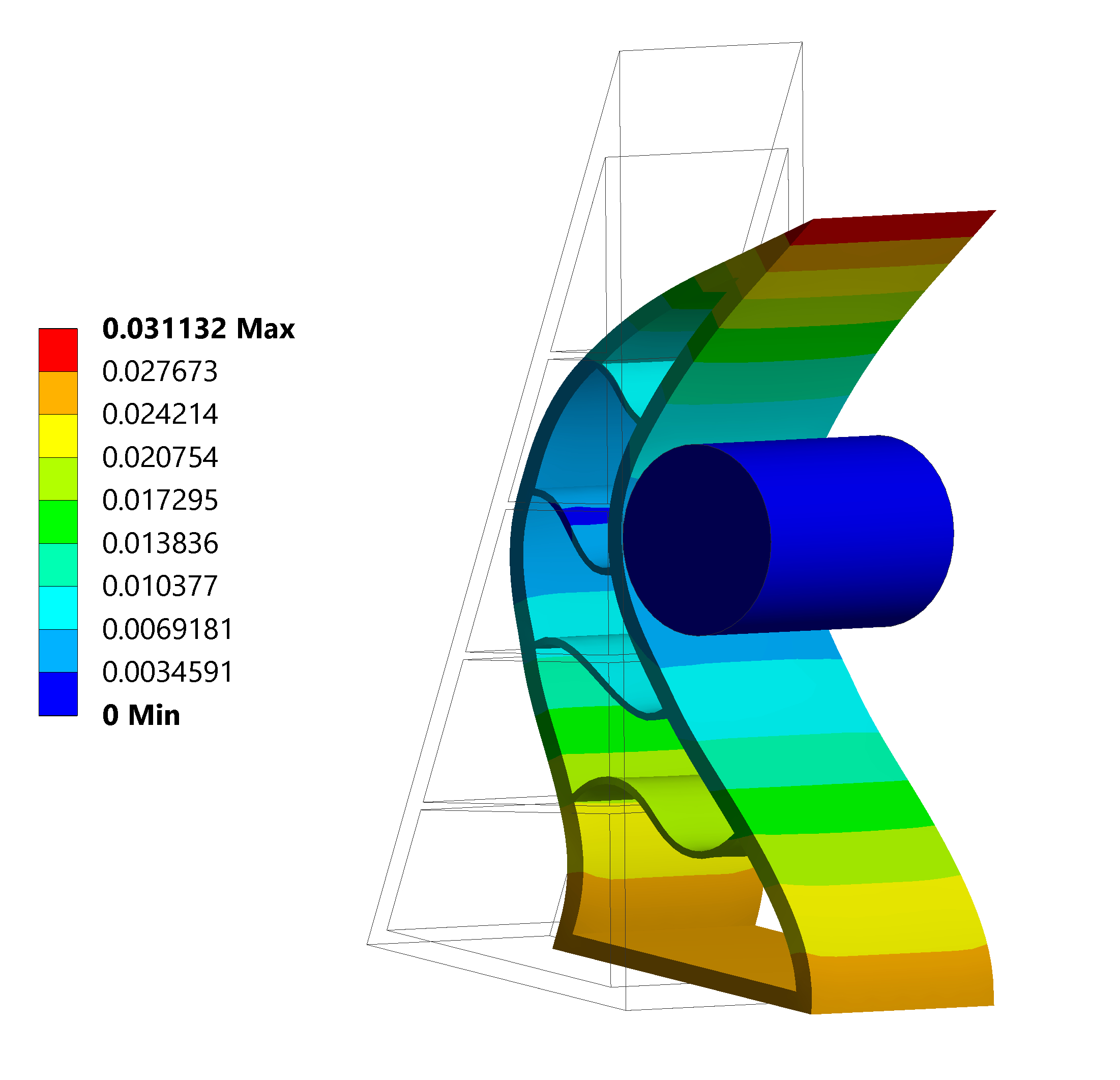}
 \caption{Total deformation contours for the specified Fin-ray finger in meters}
 \label{fig1}
\end{figure}

The total displacement contours for one of the internal structures is depicted in Fig.~\ref{fig1} as an example of the results. Parameters for this specific internal structure are front and support beam with thickness of 2 mm, the crossbeams with the thickness of 0.8 mm, and each crossbeam equal spacing of 16 mm. In addition to the deformation contour shown in Fig.~\ref{fig1}, we calculate the corresponding strains using the FEM simulations. As a result, the maximum strain in all of the scenarios is small. Therefore, TPU 95A's behavior can be approximated to behave linearly elastic in that range  \cite{Yao_Fang_Li_2023,reppel2018experimental}. 

\begin{figure}[H]
 \centering
 \includegraphics[scale=0.65]{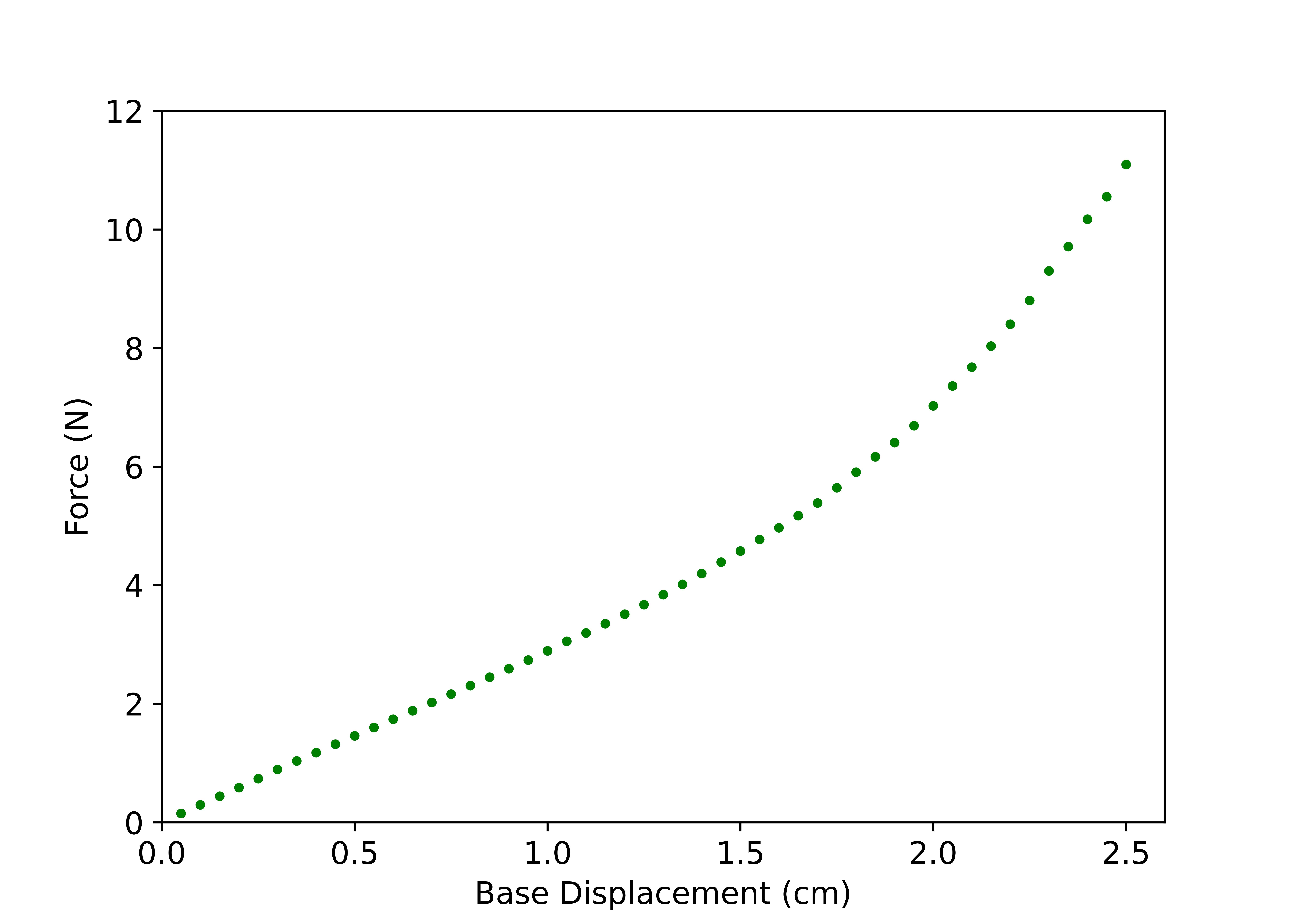}
 \caption{Contact force \textit{vs.} base displacement for the specified Fin-ray finger}
 \label{fig2}
\end{figure}

After each FEM simulation, we record values of the maximum contact force and the maximum tip displacement in both \textit{x}- and \textit{y}- directions to form a dataset. This dataset is used to train the MLP model in the next phase of the investigation. We note that maximum values of contact force and maximum tip displacement are always achieved at the maximum base displacement of the Fin-Ray finger, which is 2.5 cm. The variation of the contact force with the base displacement is displayed in Fig.~\ref{fig2} for the internal structure with front and support beam thicknesses of 2 mm, the crossbeams thickness of 0.8 mm, and each crossbeam equal spacing of 16 mm. For this proposed example, we can see that the force-displacement behavior is almost linear for the first 1.5 cm of base displacement, and in the last 1 cm of base displacement this behavior is nonlinear due to geometric and contact nonlinearities.

\section{MLP Model}
In this section, we derive the MLP model to approximate the behavior of our complex system. This model will be employed later to determine the internal structure of the Fin-Ray finger. The Min-Max feature scaling is chosen here as the normalization method. We use the Pearson correlation heatmap as a graphical tool that displays the correlation between our variables, given by Eq.~\ref{equ}. 

\begin{equation}
{r} = {\frac{\sum\left(x_{i}-\bar{x}\right)\left(y_{i}-\bar{y}\right)}{\sqrt{\sum\left(x_{i}-\bar{x}\right)^{2} \sum\left(y_{i}-\bar{y}\right)^{2}}}}
\label{equ}
\end{equation}

where $r$ is the correlation coefficient, $x_i$ are the values of the variable $x$ in the sample, and $\bar{x_i}$ is the mean of $x$. Also, $y_i$ denotes values of the variable $y$ in the sample, and $\bar{y_i}$ is the mean of $y$ \cite{Chen_2021}. This heatmap approach suggests that the correlations between each input and target feature are significant, which shows the suitability of the selected features. Other exploratory data analysis (EDA) techniques are used to reduce the effect of any outlier and address similar problems. Hence, we organize the data to ensure accuracy, consistency, and suitability for the task in a proper manner.

A multilayer perceptron is chosen to handle complex nonlinear behavior between the input and output features using intermediate hidden layers. In this feedforward neural network, the mean squared error is considered as the loss for the fitted model. The input layer of the MLP has three neurons consisting of three features: the thickness of the front and support beams, the thickness of the crossbeams, and the equal spacing between each crossbeam. The output layer consists of four neurons corresponding to the maximum contact forces and tip displacements in the \textit{x} and \textit{y}-directions, which are \( F_x \), \( F_y \), \( D_x \), and \( D_y \). The MLP has three hidden and dropout layers. We use 80\% of the data for training, 10\% for validation and the remaining 10\% for testing. Using this search technique, we tried to find the best combination of hyperparameters. Hence, the number of neurons in each hidden layer and the type of activation function for hidden layers are hyperparameters that are searched over the specified values as summarized in Table 2.
\setlength{\tabcolsep}{3pt}
\renewcommand{\arraystretch}{1.1}

\begin{table}[H]
    \begin{center}
        \caption{Grid search over specified parameter values}
        \label{tab1}
        \resizebox{0.8\columnwidth}{!}{%
        \small
        \begin{tabular}{|c | c |}
            \hline
            \textbf{\textit{Parameter Name}} & \textbf{\textit{Values}}\\
            \hline
            Number of neurons in the first hidden layer & 1-2-3-4-5-6-7-8-9-10 \\
            \hline
            Number of neurons in the second hidden layer & 1-2-3-4-5-6-7-8-9-10 \\
            \hline
            Number of neurons in the third hidden layer & 1-2-3-4-5-6-7-8-9-10 \\
            \hline
            Hidden layers activation function types & ReLU-Sigmoid-Tanh \\
            \hline
        \end{tabular}%
        }
    \end{center}
\end{table}

After the grid search, we trained the final MLP architecture with the Adam optimizer using a learning rate of 0.001, batch size of 1, and 50 epochs. Furthermore, K-Fold cross-validation is used to tackle possible overfitting problems. This approach yields lower validation errors and smoother Pareto search.

\begin{figure}[H]
    \centering
    \includegraphics[scale=0.65]{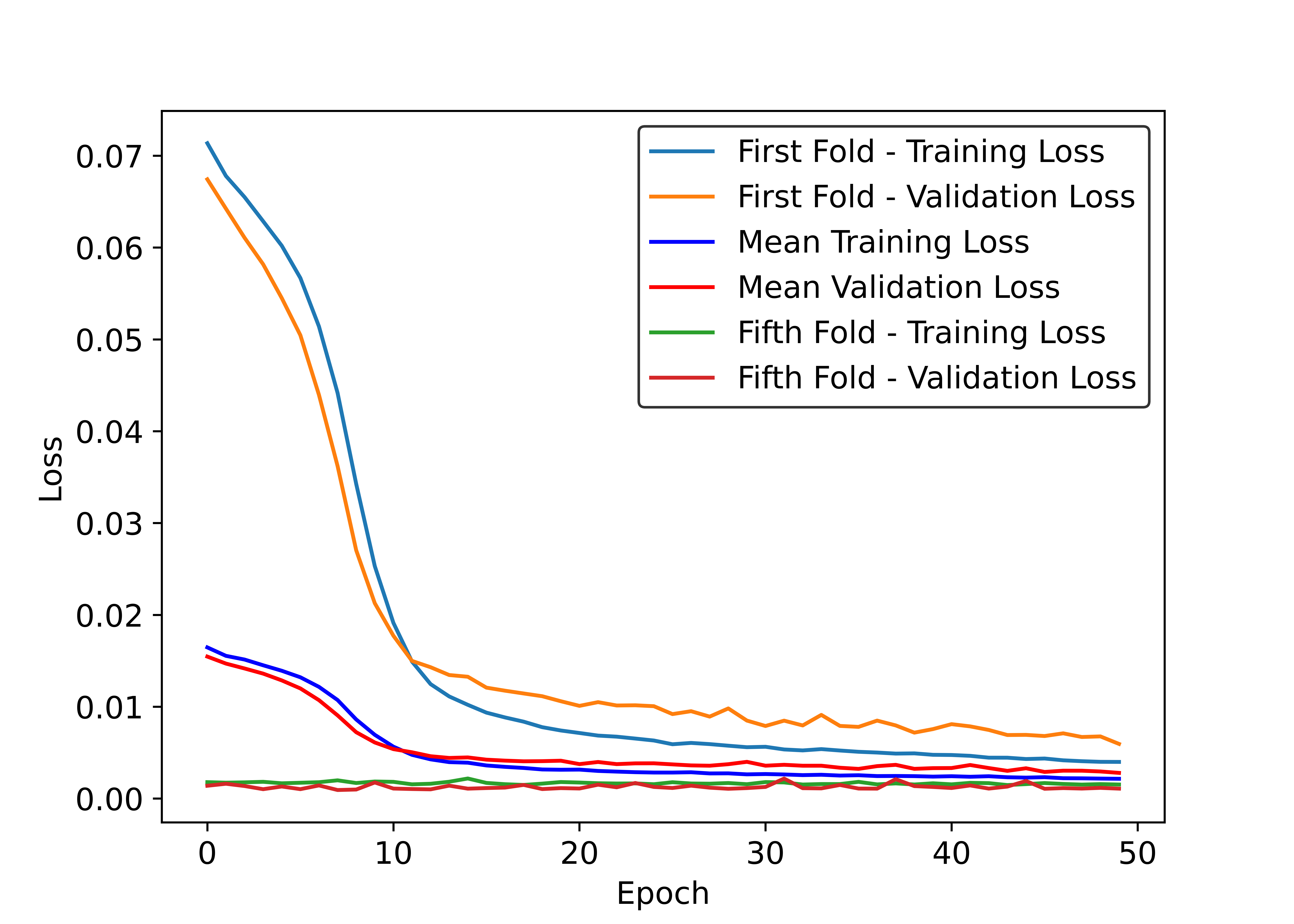}
    \caption{Training and validation loss curves}
    \label{fig3}
\end{figure}

In this search, a specific parameter value has higher scores when it performs better in the validation set. Moreover, the output layer's activation function is set to the sigmoid function to hard-encode the physics that all design variables are bounded and non-negative.. The final MLP architecture and the selected hyperparameters are summarized in Table 3. Subsequently, K-Fold cross-validation is used to ensure that our model is trained and tested on representative samples, reducing bias and enhancing the overall performance.

\setlength{\tabcolsep}{3pt}
\renewcommand{\arraystretch}{1.1}
\begin{table}[H]
    \begin{center}
        \caption{Final MLP architecture}
        \label{tab1}
        \resizebox{0.8\columnwidth}{!}{%
        \small
        \begin{tabular}{|c | c |}
            \hline
            \textbf{\textit{Parameter Name}}& \textbf{\textit{Highest Score}}\\
            \hline
            Number of neurons in the first hidden layer& 9 \\
            \hline
            Number of neurons in the second hidden layer& 10 \\
            \hline
            Number of neurons in the third hidden layer& 9 \\
            \hline
            Hidden layers activation function type& ReLU \\
            \hline
        \end{tabular}%
        }
    \end{center}
\end{table}

The loss curves for the first, last, and mean of all folds for the training and validation data decay as expected to a point of stability with a minimal gap between the two final loss values as shown in Fig.~\ref{fig3}.

The mean squared error (MSE), mean absolute error (MAE) and $R^2$ score performance metrics for training, validation, and test data demonstrate the prediction performance of the model. These metrics are given by Eqs.~\ref{eq}-\ref{eq2}.

\begin{equation}
{MAE} = \frac{1}{N}{\sum_{i=1}^{N}}\left| y_{i}-\hat{y} \right|
\label{eq}
\end{equation}

\begin{equation}
{MSE} = {\frac{1}{N}{\sum_{i=1}^{N}}(y_{i}-\hat{y})^2}
\label{eq1}
\end{equation}

\begin{equation}
{R^2} = {1 - \frac{{\sum_{i=1}^{N}}(y_{i}-\hat{y})^2}{{\sum_{i=1}^{N}}(y_{i}-\bar{y})^2}}
\label{eq2}
\end{equation}

where $N$ is the number of data points, $\hat{y}$ indicates the predicted value of $y$, $y_i$ is the values of $y$ variable in the sample, and $\bar{y}$ denotes the mean value of $y$. The mean squared error (MSE), mean absolute error (MAE) and $R^2$ score for test data of the target features are tabulated in Table 4. We note that features $F_{x}$ and $F_{y}$, appearing on the table, represent the contact forces in the \textit{x}- and \textit{y}-directions, respectively. In a similar vein, features $D_{x}$ and $D_{y}$ indicate the tip displacements, respectively, in the \textit{x}- and \textit{y}- directions. These numbers suggest that the model is suitably generalized to unseen data and our model has a balance between bias and variance, avoiding overfitting. Therefore, based on the performance metrics and numerical results presented here, the MLP is capable of predicting of the target features for various internal structures.

\setlength{\tabcolsep}{3pt}
\renewcommand{\arraystretch}{1.7}
\begin{table}[H]
    \centering
    \caption{MSE, MAE, and $R^2$ score performance metrics for test set}
    \label{tab1}
    \resizebox{0.6\columnwidth}{!}{%
    \small
    \begin{tabular}{|c|c|c|c|}
        \hline
        \textbf{\textit{Feature}} & \textbf{\textit{MSE}} & \textbf{\textit{MAE}} & \textbf{\textit{$R^2$ score}} \\
        \hline
        F\textsubscript{x} & 0.00055 & 0.01823 & 0.99 \\
        \hline
        F\textsubscript{y} & 0.00350 & 0.04569 & 0.96 \\
        \hline
        D\textsubscript{x} & 0.00320 & 0.04958 & 0.94 \\
        \hline
        D\textsubscript{y} & 0.00152 & 0.03419 & 0.97 \\
        \hline
    \end{tabular}
    }
\end{table}

\section{Multi-objective Design Optimization}
We employ a multi-objective optimization algorithm to find the internal structure for the Fin-Ray finger. We use NSGA-II, which generates offspring by using a particular kind of crossover and mutation method. The selection process for the next generations is based on comparisons of the non-dominated sorting and crowding distances \cite{996017}. Three design parameters are considered in the dimensional optimization, including the thickness of the front and support beams, the thickness of the crossbeams, and the equal spacing between the crossbeams, which are considered input features for the MLP in the previous section. These design parameters are constrained to the range in which the MLP is trained, as given in Table 1. The MLP model developed in the previous section is used here for accurate prediction of our objectives. Two optimization objectives are readily determined from the four target features of the MLP:

\begin{equation}
{F} = {\sqrt{F_{x}^2+{F_{y}^2}}}
\label{eq3}
\end{equation}

\begin{equation}
{D} = {\sqrt{D_{x}^2+{D_{y}^2}}}
\label{eq4}
\end{equation}

where $F$ and $D$ represent the predicted magnitudes of the maximum contact force and magnitude of the maximum tip displacement. Also, $F_{x}$ and $D_{x}$ are, respectively, the predicted maximum force and maximum tip displacement in the \textit{x}-direction, and $F_{y}$ and $D_{y}$ are the corresponding quantities in the \textit{y}-direction. As aforementioned, these magnitudes are considered as our two objectives. The maximum tip displacement towards the grasped object is regarded a measure for how suitable a design is at handling delicate objects and highlights the Fin-Ray finger's adaptability to the cylindrical objects. A higher value for the contact force indicates the tendency of the finger to possess a higher load capacity. Achieving a large grasping force demands enhanced rigidity, which is in contrast with the first objective and it is opposed to the softness necessary for delicate grasping.

After the determination of the objectives and design parameters, NSGA-II is utilized to find the design of the Fin-Ray finger. The Pymoo library allows us to implement the NSGA-II algorithm by defining our problem with the specific parameters mentioned in Table 5 \cite{pymoo}. We choose population size of 500 to maintain diversity, and we give sufficient evolution depth without excessive cost using 100 generations. The optimization termination criterion is considered as a maximum of 50 generations. No specific optimization constraint is defined for the optimization problem. The three input variables are normalized between 0 and 1, and their real bounds are limited to the geometric specifications. A crossover rate of 0.9 pushes mixing, while a mutation rate of 0.1 brings diversity without over-perturbing designs. These numbers are aligned with common practice and lead to satisfactory results here. For other optimization variables, parameters, and methods, the Pymoo default setup is used.

\begin{figure}[H]
 \centering
 \includegraphics[scale=0.65]{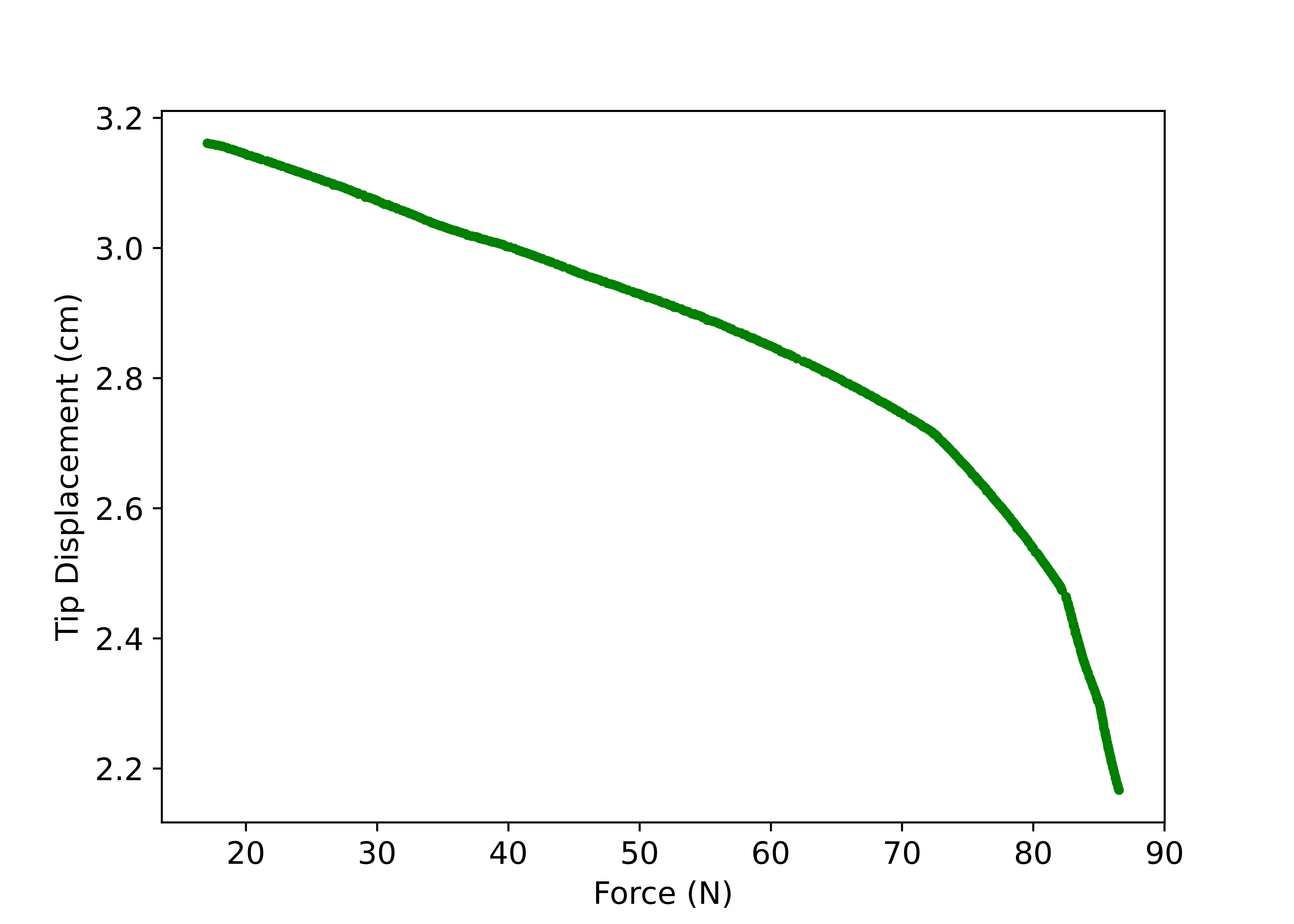}
 \caption{The set of all Pareto efficient solutions}
 \label{fig4}
\end{figure}

\setlength{\tabcolsep}{3pt}
\renewcommand{\arraystretch}{1.1}
\begin{table}[H]
    \begin{center}
        \caption{NSGA-II algorithm parameters}
        \label{tab1}
        \resizebox{0.4\columnwidth}{!}{%
        \small
            \begin{tabular}{|c | c |}
            \hline
            \textbf{\textit{Parameter Name}}& \textbf{\textit{Value}}\\
            \hline
            Number of generations& 100 \\
            \hline
            Population size& 500 \\
            \hline
            Crossover rate& 0.9 \\
            \hline
            Mutation rate& 0.1 \\
            \hline
        \end{tabular}%
        }
    \end{center}
\end{table}
The set of optimal solutions, which is also known as the Pareto front, is presented in Fig.~\ref{fig4}. The Pareto front narrows down the solutions to a set in which we can make a choice in the trade-off between objectives. Depending on the application, one of these solutions is selected to achieve a desirable design.

\begin{figure}[!h]
 \centering
 \includegraphics[scale=0.65]{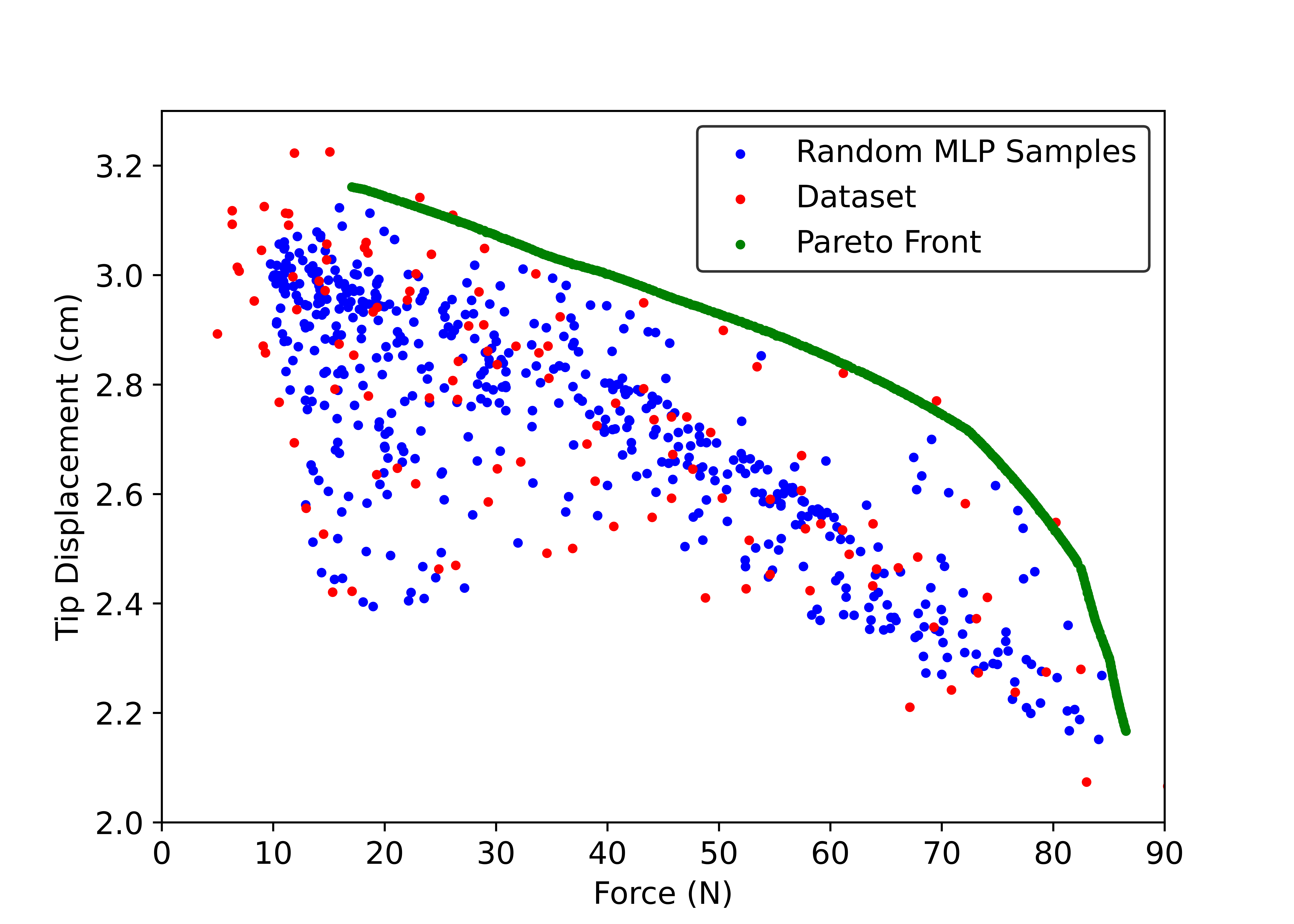}
 \caption{Random MLP Samples, dataset, and Pareto front}
 \label{fig5}
\end{figure}

We also generate a series of random points, feed them into the MLP, and compare the predicted target features with the Pareto front. As illustrated in Fig.~\ref{fig5}, the Pareto front is not dominated by these synthetic samples. With the limited amount of training data available, the MLP showed a slight tendency to underestimate the Fin-Ray forces and displacements compared to the dataset points, particularly near the regions of the Pareto front. However, it mainly reproduced nearby trade-offs. This effect can be attributed to the regression toward the mean, which can be observed in neural network models trained with limited data. Consequently, the predicted forces and displacements along the Pareto front are generally lower than those obtained from FEM simulations for the same geometries.

In order to examine the Pareto front results, we select three points from it and compare them with the FEM simulation results. These three points and their corresponding Fin-Ray finger geometric parameters are presented in Table 6. Depending on the intended application of the Fin-Ray gripper, the chosen design may vary. Here, Points A and B represent the designs with the highest tip displacement and the highest force, respectively. In addition to these points, Point C is considered a balanced solution. Since the displacement and force values along the Pareto front have different ranges, these values were normalized between 0 and 1. Among the normalized Pareto front points, the one with the minimum Euclidean distance to the (1,1) is selected as Point C. The FEM simulation results, Pareto front data, and corresponding errors are also summarized in Table 7. It is expected to achieve higher contact forces when the overall stiffness of the Fin-Ray structure increases. Relatively, this can be addressed when the thickness of different parts of the Fin-Ray gets larger. In this case, the resulting Fin-Ray would not be suitable for delicate tasks. However, according to Table 6, it is shown that font and support beam thicknesses have larger impact on the objectives, while crossbeam thickness is less impactful. Regarding tip displacement, this comparison demonstrates a relatively low error between the FEM and Pareto front outputs. The error increases for the force column. This behavior is somehow expected because forces are more sensitive to local stiffness changes and nonlinear effects than tip displacements and small differences in the predicted deformation can lead to more noticeable differences in the reaction forces. On the other hand, given the limited FEM dataset, the predictive model naturally prioritizes the smoother and dominant trends in the input and output mapping. Tip displacements vary relatively smoothly with the geometric parameters, whereas contact forces exhibit stronger nonlinearities, especially in the high-force region as depicted in Fig.~\ref{fig2}. Therefore, the predictive model makes lower relative errors for displacements compared to forces. Nevertheless, these errors remain within an acceptable range for the design parameters and trade-off analysis targeting in this work.

\begin{table}[H]
    \centering
    \small
    \caption{Geometry properties of selected Pareto front structures}
    \label{tab:geometry_properties}
    \resizebox{\textwidth}{!}{%
    \small
    \begin{tabular}{|c|c|c|c|}
        \hline
        \textbf{\textit{Point}} & \textbf{\textit{Front and Support}} & \textbf{\textit{Crossbeams}} & \textbf{\textit{Spacing Distance}} \\
        & \textbf{\textit{Beams Thicknesses (mm)}} & \textbf{\textit{Thickness (mm)}} & \textbf{\textit{(mm)}} \\
        \hline
        A & 1.500 & 1.600 & 10.000 \\
        \hline
        B & 4.000 & 1.600 & 11.320 \\
        \hline
        C & 3.310 & 1.599 & 10.000 \\
        \hline
    \end{tabular}%
    }
\end{table}

\begin{table}[H]
    \centering
    
    \caption{Comparison of Pareto front and FEM simulations results}
    \label{tab:results_comparison}
    \resizebox{\textwidth}{!}{%
    \small
    \begin{tabular}{|c|c|c|c|c|c|c|}
        \hline
        \textbf{\textit{Point}} & \multicolumn{2}{c|}{\textbf{\textit{Pareto Front Results}}} & \multicolumn{2}{c|}{\textbf{\textit{FEM Results}}} & \multicolumn{2}{c|}{\textbf{\textit{Error (\%)}}} \\
        \cline{2-7}
        & \textbf{\textit{Displacement}} & \textbf{\textit{Force}} & \textbf{\textit{Displacement}} & \textbf{\textit{Force}} & \textbf{\textit{Displacement}} & \textbf{\textit{Force}} \\
        \hline
        A& 31.609 mm & 17.065 N & 33.223 mm & 16.032 N & 4.857 & 6.443 \\
        \hline
        B& 21.668 mm & 86.536 N & 21.837 mm & 94.143 N & 0.777 & 8.080 \\
        \hline
        C& 27.845 mm & 66.583 N & 28.196 mm & 61.276 N & 1.246 & 8.661 \\
        \hline
    \end{tabular}%
    }
\end{table}

\section{Conclusion}
\label{sec1}
In this research, we have presented a comprehensive approach for designing Fin-Ray fingers to enhance grasping performance. A finite element framework for predicting the forces during the finger's interaction with cylindrical objects has been proposed. Using finite element simulations, we have developed a dataset consisting of 120 simulation results of the tip displacements and the contact forces for distinct Fin-ray internal structures. A model has been forged through the training of the neural network. The MLP hyperparameters have been selected by a grid search over the specified range of values. This MLP model has been used to find the designs with the NSGA-II algorithm. The set of all Pareto efficient solutions has been found and validated with the random data generated by the MLP. Despite differences in the predicted force and displacement values by the MLP model compared to the simulation output of the same geometry, the optimization framework identified geometries with respect to the defined objectives. This shows that the surrogate model is sufficiently accurate to produce valid designs and show trade-offs. To assess the results of the optimization process, three different points from the Pareto front have been selected, and using their geometric properties, we have simulated the corresponding Fin-Ray finger structures. According to the simulation results of these fingers, we have compared the results to the Pareto front data. This comparison shows a promising relation between FEM simulations and design optimization solutions. However, performing experiments on selected Fin-Ray finger structures is suggested as future studies to validate the results. In addition, in this study we have considered the magnitude of tip displacement and contact force as the design objectives to enhance grasping performance. Moreover, our simulations have been done with only a certain geometry for the grasped object. In addition, future work should focus on dataset improvements to increase surrogate reliability. As a future study, our methodology can be generalized to more complicated geometries, better representatives for design objectives, different object sizes and materials as well. In conclusion, it is found that the present methodology achieved the desired results for designing the Fin-ray finger, assisting in making a choice between objectives.



\bibliographystyle{elsarticle-num} 
\bibliography{newreferences}

@ARTICLE{Xin2023TheRO,
author = {Xin, Yangyang and Zhou, Xinran and Bark, Hyunwoo and Lee, Pooi See},
title = {The Role of 3{D} Printing Technologies in Soft Grippers},
journal = {Advanced Materials},
volume = {36},
number = {34},
year = {2024},
}

@ARTICLE{996017,
  author={Deb, K. and Pratap, A. and Agarwal, S. and Meyarivan, T.},
  journal={IEEE Transactions on Evolutionary Computation}, 
  title={A fast and elitist multiobjective genetic algorithm: NSGA-{II}}, 
  year={2002},
  volume={6},
  number={2},
  pages={182-197},
}

@ARTICLE{pymoo,
    author={J. {Blank} and K. {Deb}},
    journal={IEEE Access},
    title={pymoo: Multi-Objective Optimization in Python},
    year={2020},
    volume={8},
    number={},
    pages={89497-89509},
}

@INPROCEEDINGS{10903660,
  author={Ghanizadeh, Ali and Ghanbili, Farshad and Bahrami, Arash},
  booktitle={2024 12th RSI International Conference on Robotics and Mechatronics (ICRoM)}, 
  title={Contact Force Estimation and Control in Soft Fin-Ray Grippers Using {FEM} and Experimental Approach}, 
  year={2024},
  volume={},
  number={},
  pages={076-082},
  }

@ARTICLE{article,
author = {Shintake, Jun and Cacucciolo, Vito and Floreano, Dario and Shea, Herbert},
year = {2018},
month = {05},
title = {Soft Robotic Grippers},
volume = {30},
journal = {Advanced Materials},
}

@Article{pfaff2011application,
  title={Application of fin ray effect approach for production process automation},
  author={Pfaff, Ondrej and Simeonov, Simeon and Cirovic, Ivan and Stano, Pavol},
  journal={Annals \& Proceedings of DAAAM International},
  volume={22},
  number={1},
  pages={1247--1249},
  year={2011},
  publisher={DAAAM International Vienna},
  
}

@article{Jin_2025,
year = {2025},
month = {02},
publisher = {IOP Publishing},
volume = {34},
number = {3},
author = {Jin, Liuchao and Zhai, Xiaoya and Xue, Wenbo and Zhang, Kang and Jiang, Jingchao and Bodaghi, Mahdi and Liao, Wei-Hsin},
title = {Finite element analysis, machine learning, and digital twins for soft robots: state-of-arts and perspectives},
journal = {Smart Materials and Structures},
}

@ARTICLE{10.3389/frobt.2016.00070,
AUTHOR={Crooks, Whitney  and Vukasin, Gabrielle  and O’Sullivan, Maeve  and Messner, William  and Rogers, Chris },
TITLE={Fin Ray Effect Inspired Soft Robotic Gripper: From the RoboSoft Grand Challenge toward Optimization},       
JOURNAL={Frontiers in Robotics and AI},      
VOLUME={3},
YEAR={2016},
ISSN={2296-9144},
}

@INPROCEEDINGS{8813388,
  author={Ali, Md. Hazrat and Zhanabayev, Asset and Khamzhin, Samat and Mussin, Kainar},
  booktitle={2019 5th International Conference on Control, Automation and Robotics (ICCAR)}, 
  title={Biologically Inspired Gripper Based on the Fin Ray Effect}, 
  year={2019},
  pages={865-869},
}

@ARTICLE{doi:10.1177/0278364920913926,
author = {Xiaowei Shan and Lionel Birglen},
title ={Modeling and analysis of soft robotic fingers using the fin ray effect},
journal = {The International Journal of Robotics Research},
volume = {39},
number = {14},
pages = {1686-1705},
year = {2020},
}

@INPROCEEDINGS{9115969,
  author={Elgeneidy, Khaled and Fansa, Adel and Hussain, Irfan and Goher, Khaled},
  booktitle={2020 3rd IEEE International Conference on Soft Robotics (RoboSoft)}, 
  title={Structural Optimization of Adaptive Soft Fin Ray Fingers with Variable Stiffening Capability}, 
  year={2020},
  pages={779-784},
}

@ARTICLE{9380238,
  author={Xu, Wenfu and Zhang, Heng and Yuan, Han and Liang, Bin},
  journal={IEEE Transactions on Robotics}, 
  title={A Compliant Adaptive Gripper and Its Intrinsic Force Sensing Method}, 
  year={2021},
  volume={37},
  number={5},
  pages={1584-1603},
}

@ARTICLE{10.3389/frobt.2021.631371,  
AUTHOR={De Barrie, Daniel  and Pandya, Manjari  and Pandya, Harit  and Hanheide, Marc  and Elgeneidy, Khaled },         
TITLE={A Deep Learning Method for Vision Based Force Prediction of a Soft Fin Ray Gripper Using Simulation Data},
JOURNAL={Frontiers in Robotics and AI},   
VOLUME={8},
YEAR={2021},
}

@ARTICLE{10.3389/frobt.2020.590076,
AUTHOR={Deng, Zhifeng  and Li, Miao },       
TITLE={Learning Optimal Fin-Ray Finger Design for Soft Grasping},      
JOURNAL={Frontiers in Robotics and AI},       
VOLUME={7},
YEAR={2021},
}

@ARTICLE{YAO2024105472,
title = {Design optimization of soft robotic fingers biologically inspired by the fin ray effect with intrinsic force sensing},
journal = {Mechanism and Machine Theory},
volume = {191},
year = {2024},
author = {Jiaqiang Yao and Yuefa Fang and Xinhua Yang and Peiyi Wang and Luquan Li},
}

@ARTICLE{Yao_Fang_Li_2023,
title={Research on effects of different internal structures on the grasping performance of Fin Ray soft grippers},
volume={41},
number={6},
journal={Robotica},
author={Yao, Jiaqiang and Fang, Yuefa and Li, Luquan}, year={2023},
pages={1762–1777},
}

@ARTICLE{article12,
author = {Gidugu, Lakshmi and Javed, Arshad and Faller, Lisa-Marie},
year = {2024},
month = {05},
title = {Versatile 3{D}-printed fin-ray effect soft robotic fingers: lightweight optimization and performance analysis},
volume = {46},
journal = {Journal of the Brazilian Society of Mechanical Sciences and Engineering},
}

@INPROCEEDINGS{10802217,
  author={Wang, Xing and Dabrowski, Joel Janek and Pinskier, Josh and Liow, Lois and Viswanathan, Vinoth and Scalzo, Richard and Howard, David},
  booktitle={2024 IEEE/RSJ International Conference on Intelligent Robots and Systems (IROS)}, 
  title={{PINN}-Ray: A Physics-Informed Neural Network to Model Soft Robotic Fin Ray Fingers}, 
  year={2024},
  volume={},
  number={},
  pages={247-254},
}

@ARTICLE{app11093858,
AUTHOR = {Suder, Jiří and Bobovský, Zdenko and Mlotek, Jakub and Vocetka, Michal and Oščádal, Petr and Zeman, Zdeněk},
TITLE = {Structural Optimization Method of a FinRay Finger for the Best Wrapping of Object},
JOURNAL = {Applied Sciences},
VOLUME = {11},
YEAR = {2021},
NUMBER = {9},
ARTICLE-NUMBER = {3858},
}

@ARTICLE{AN2025110118,
title = {Linkage integrated fin ray gripper capable of safe adaptive grasping for tomato harvesting},
journal = {Computers and Electronics in Agriculture},
volume = {232},
pages = {110-118},
year = {2025},
author = {Byeongchan An and Taeyong Choi and Uikyum Kim},
}

@ARTICLE{Chen_2021,
year = {2021},
month = {01},
publisher = {IOP Publishing},
volume = {1757},
number = {1},
author = {Chen, Pengtian and Li, Fei and Wu, Chunwang},
title = {Research on Intrusion Detection Method Based on Pearson Correlation Coefficient Feature Selection Algorithm},
journal = {Journal of Physics: Conference Series},
}

@article{reppel2018experimental,
  title={Experimental determination of elastic and rupture properties of printed Ninjaflex},
  author={Reppel, Thomasa and Weinberg, Kerstin},
  journal={Technische Mechanik-European Journal of Engineering Mechanics},
  volume={38},
  number={1},
  pages={104--112},
  year={2018},
}







\end{document}